\documentclass{article}

\usepackage[preprint,nonatbib]{neurips_2022}

\usepackage[utf8]{inputenc} 
\usepackage[T1]{fontenc}    
\usepackage{hyperref}       
\usepackage{url}            
\usepackage{booktabs}       
\usepackage{amsfonts}       
\usepackage{nicefrac}       
\usepackage{microtype}      
\usepackage{xcolor}         

\usepackage{times}
\usepackage{epsfig}
\usepackage{graphicx}
\usepackage{amsmath}
\usepackage{amssymb}
\usepackage{color}
\usepackage{algorithm}
\usepackage{algorithmic}
\usepackage{multirow}
\usepackage{pifont}

\newcommand\ie{\emph{i.e.}}
\newcommand\eg{\emph{e.g.}}
\newcommand{\xmark}{\ding{56}}%

\title{Semi-supervised Vision Transformers at Scale}

\author{
Zhaowei Cai, Avinash Ravichandran, Paolo Favaro, Manchen Wang, \\ \textbf{Davide Modolo, Rahul Bhotika, Zhuowen Tu, Stefano Soatto} \\
AWS AI Labs \\
\texttt{\{zhaoweic,ravinash,pffavaro,manchenw,dmodolo,ztu,soattos\}@amazon.com} \\
}

\begin{document}

\maketitle

\begin{abstract}
  We study semi-supervised learning (SSL) for vision transformers (ViT), an under-explored topic despite the wide adoption of the ViT architectures to different tasks.
  To tackle this problem, we propose a new SSL pipeline, consisting of first \emph{un/self-supervised pre-training}, followed by \emph{supervised fine-tuning}, and finally \emph{semi-supervised fine-tuning}. At the semi-supervised fine-tuning stage, we adopt an exponential moving average (EMA)-Teacher framework instead of the popular FixMatch, since the former is more stable and delivers higher accuracy for semi-supervised vision transformers. In addition, we propose a \emph{probabilistic pseudo mixup} mechanism to interpolate unlabeled samples and their pseudo labels for improved regularization, which is important for training ViTs with weak inductive bias. Our proposed method, dubbed \emph{Semi-ViT}, achieves comparable or better performance than the CNN counterparts in the semi-supervised classification setting. Semi-ViT also enjoys the scalability benefits of ViTs that can be readily scaled up to large-size models with increasing accuracies. For example, Semi-ViT-Huge achieves an impressive 80\% top-1 accuracy on ImageNet using only 1\% labels, which is comparable with Inception-v4 using 100\% ImageNet labels.
\end{abstract}

\section{Introduction}

\begin{figure*}[h!]
\begin{minipage}[b]{.31\linewidth}
\centering
\centerline{\epsfig{figure=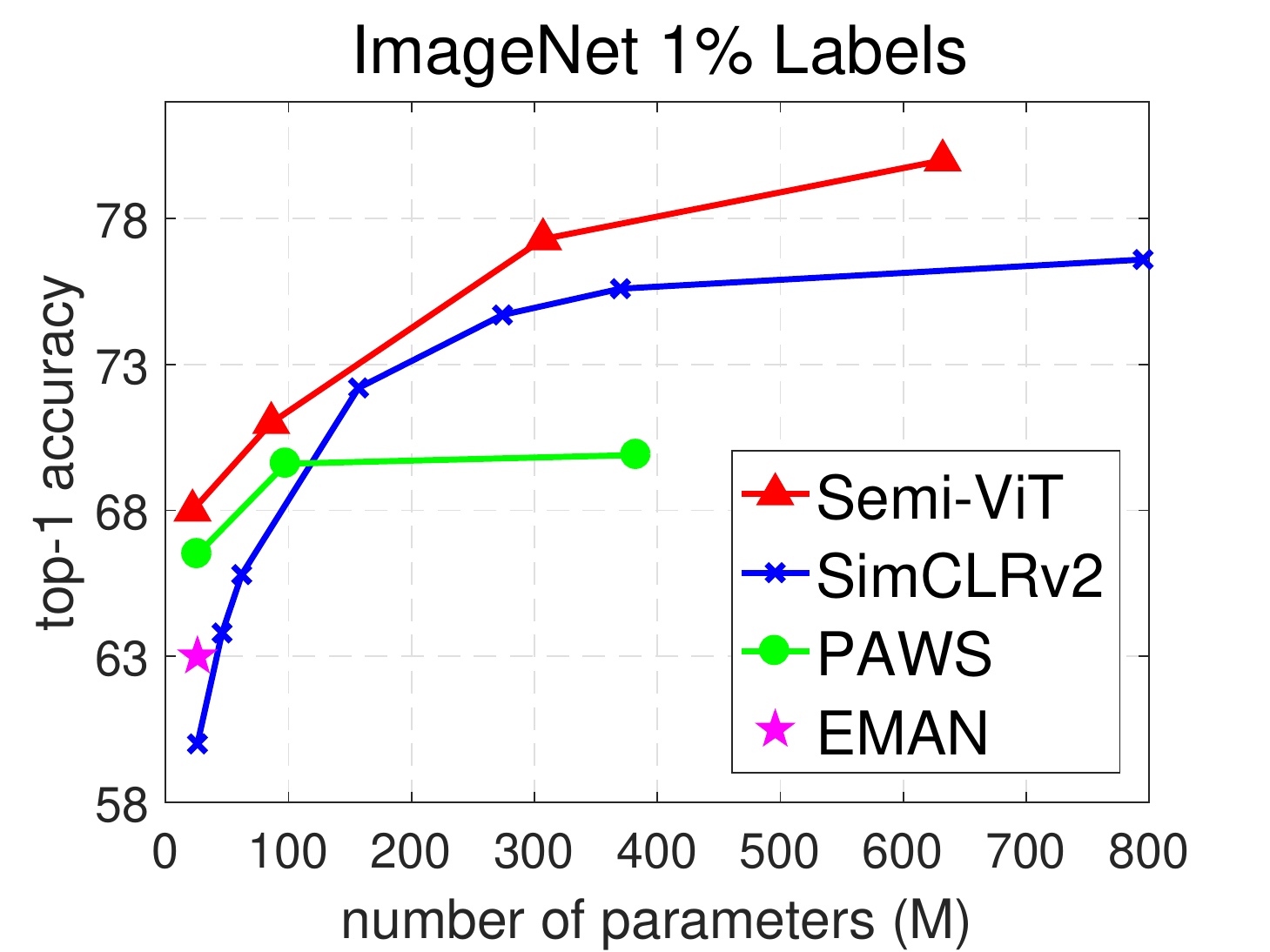,width=4.69cm,height=3.52cm}}{(a)}
\end{minipage}
\hfill
\begin{minipage}[b]{.31\linewidth}
\centering
\centerline{\epsfig{figure=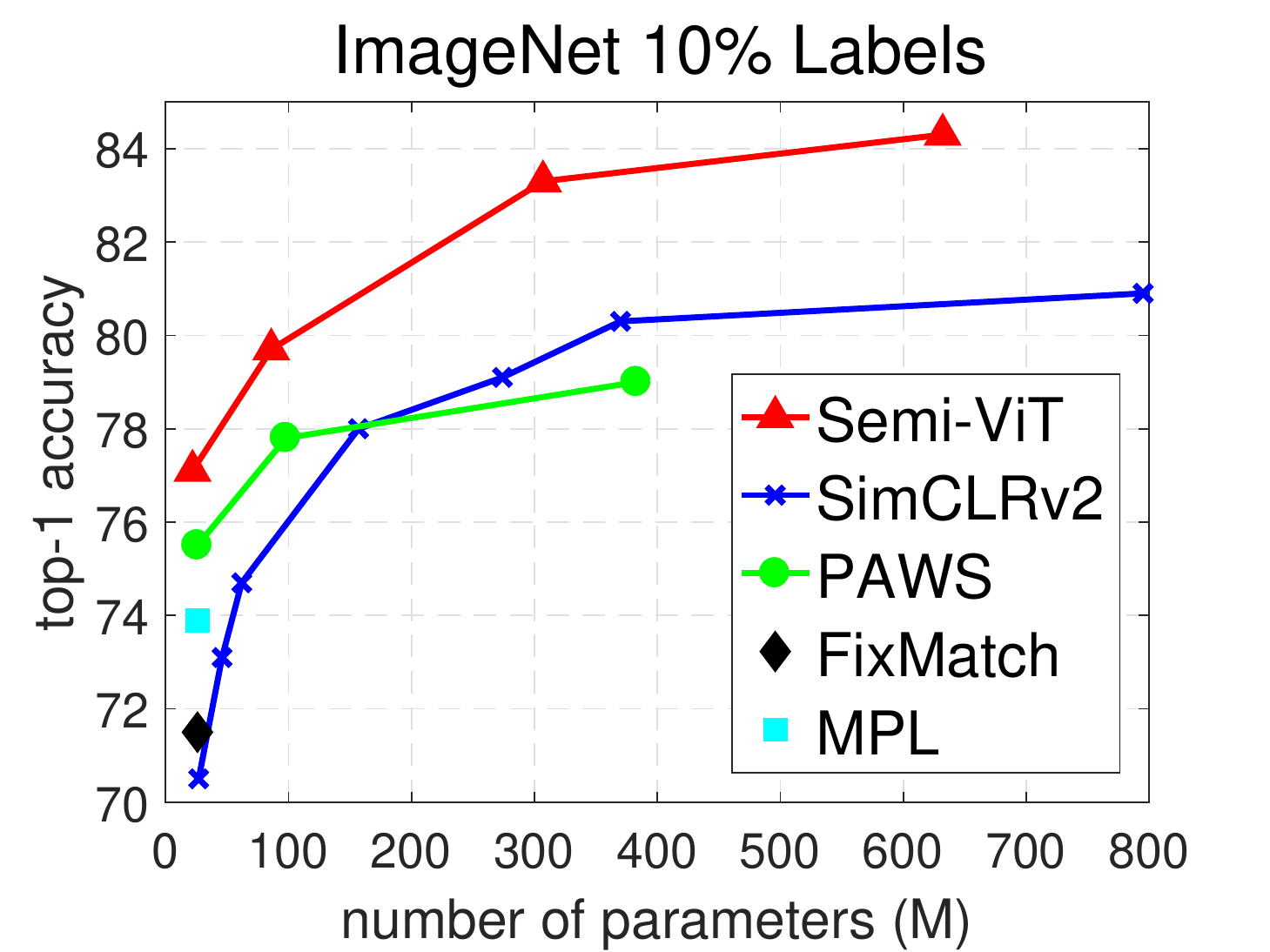,width=4.69cm,height=3.52cm}}{(b)}
\end{minipage}
\hfill
\begin{minipage}[b]{.36\linewidth}
\centering
\centerline{\epsfig{figure=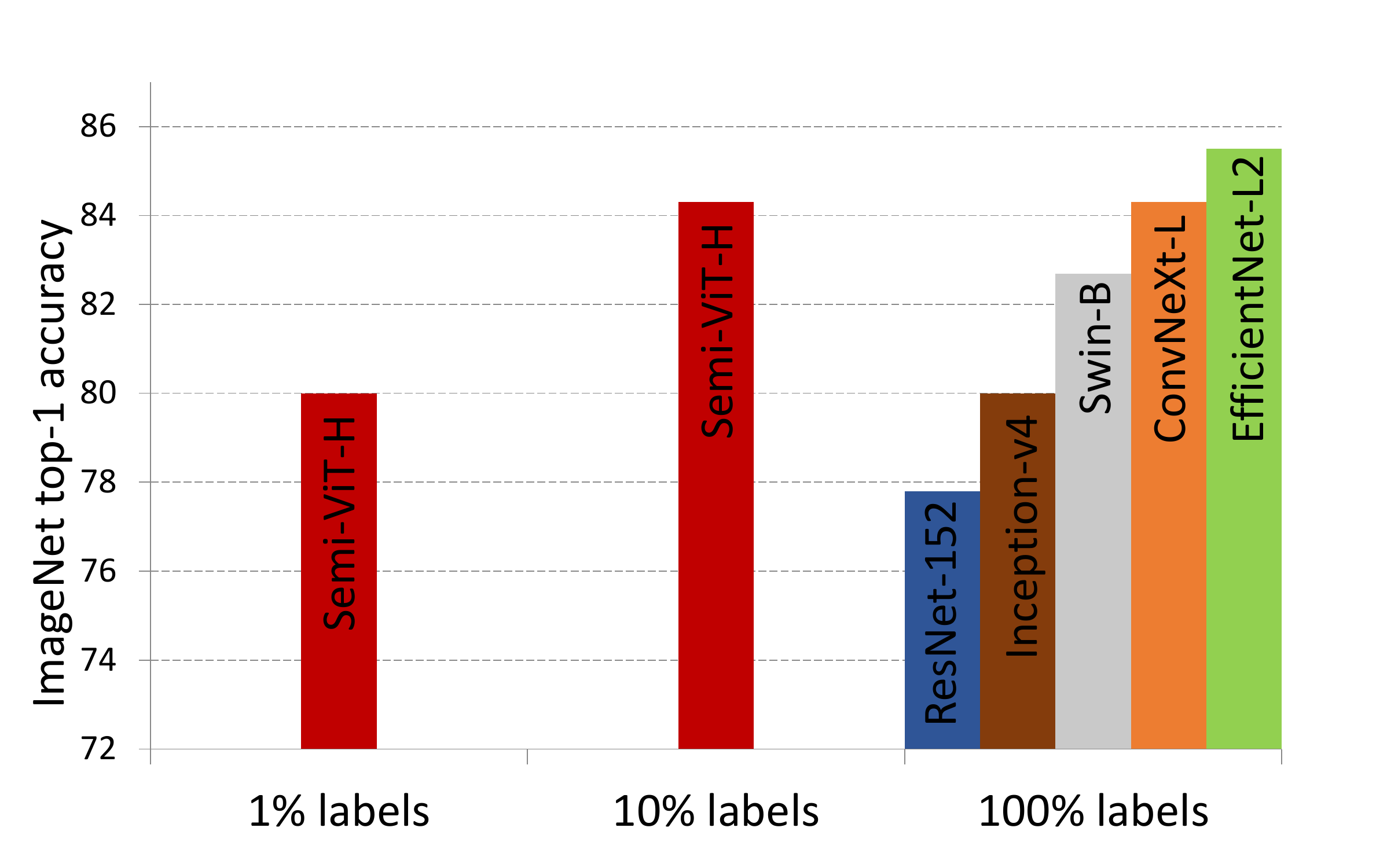,width=5.38cm,height=3.52cm}}{(c)}
\end{minipage}
\caption{(a) and (b) are the comparisons of our Semi-ViT with the state-of-the-art SSL algorithms at different model scales, and (c) is the comparison with the state-of-the-art supervised models.}
\label{fig:sota}
\end{figure*}

In the past few years, Vision Transformers (ViT) \cite{DBLP:conf/iclr/DosovitskiyB0WZ21}, which adapt the transformer architectures \cite{DBLP:conf/nips/VaswaniSPUJGKP17} to the visual domain, have achieved remarkable progresses in supervised learning \cite{DBLP:conf/icml/TouvronCDMSJ21,DBLP:conf/iccv/LiuL00W0LG21,DBLP:conf/iccv/XuXCT21}, un/self-supervised learning \cite{DBLP:conf/iccv/ChenXH21,DBLP:conf/iccv/CaronTMJMBJ21,he2021masked}, and many other computer vision tasks \cite{DBLP:conf/eccv/CarionMSUKZ20,DBLP:conf/iccv/0001XMLYMF21,DBLP:conf/iccv/Arnab0H0LS21,DBLP:conf/iccv/StrudelPLS21} (with architecture modifications).
However, ViTs have yet to show the same advantage in semi-supervised learning (SSL), where only a small subset of the training data is labeled, a problem in the middle between supervised and un/self-supervised learning.
Although several recent methods in SSL have significantly advanced the field \cite{DBLP:conf/iclr/LaineA17,DBLP:conf/nips/TarvainenV17,DBLP:conf/nips/BerthelotCGPOR19,DBLP:conf/nips/SohnBCZZRCKL20,DBLP:conf/nips/XieDHL020,DBLP:conf/cvpr/CaiRMFTS21,DBLP:conf/cvpr/PhamDXL21}, the transfer of these methods from Convolutional Neural Networks (CNN) to ViT architectures has yet to show much promise. 
For example, as discussed in \cite{weng2021semi}, the direct application of FixMatch \cite{DBLP:conf/nips/SohnBCZZRCKL20}, one of the most popular SSL methods, to ViT leads to an inferior performance (about 10 points worse) than when used with a CNN architecture. The challenge could be potentially caused by the fact that ViTs are known to require more data for training and to have a weaker inductive bias than CNNs \cite{DBLP:conf/iclr/DosovitskiyB0WZ21}. However, in this paper we show that semi-supervised ViTs can outperform the CNN counterparts when trained properly, suggesting promising potential to advance SSL beyond CNN architectures. 

To achieve such success, we propose the following SSL pipeline: 1) \textit{un/self-supervised pre-training} on all data (both labeled and unlabeled), followed by 2) \textit{supervised fine-tuning} only on labeled data, and finally 3) \textit{semi-supervised fine-tuning} on all data. This new pipeline is stable and helps reduce the sensitivity of hyperparameter tuning when training ViTs for SSL in our experiments. At the final stage of \textit{semi-supervised fine-tuning}, we adopt the EMA-Teacher framework \cite{DBLP:conf/nips/TarvainenV17,DBLP:conf/cvpr/CaiRMFTS21}, an improved version over the popular FixMatch \cite{DBLP:conf/nips/SohnBCZZRCKL20}. Unlike FixMatch that often fails to converge when training semi-supervised ViT, EMA-Teacher shows more stable training behaviors and better performance. 
In addition, we propose \emph{probabilistic pseudo mixup} for the pseudo-labeling based SSL methods, which interpolates the unlabeled samples coupled with pseudo labels for enhanced regularization. In the standard mixup \cite{DBLP:conf/iclr/ZhangCDL18} the mixup ratio is randomly sampled from a Beta distribution. In contrast, in the \emph{probabilistic pseudo mixup} the ratio depends on the respective confidences of two mixed-up samples, such that the sample with higher confidence will weigh more in the final interpolated sample. This new data augmentation technique brings non-negligible gains since ViT has weak inductive bias, especially for scenarios where the training is more difficult, \eg, without un/self-supervised pre-training or on data regimes with very few labeled samples (\eg, 1\% labels). We call our method \emph{Semi-ViT}. Notice that Semi-ViT is built on exactly the same design of ViTs (\ie, there are neither additional parameters nor architectural changes).  

Semi-ViT achieves promising results from different aspects (Figure \ref{fig:sota}).
1) For the first time, we show that \emph{pure} ViTs can reach comparable or better accuracy than CNNs on SSL\footnote{Although \cite{weng2021semi} was the first to use transformer for SSL, it is a mixture architecture of CNN and ViT and requires to use CNN as the teacher to produce pseudo labels.}.
2) Semi-ViT can be readily scaled up under the SSL setting. 
This is illustrated in Figure~\ref{fig:sota}~(a) and (b) on ViT architectures at different scales, ranging from ViT-Small to ViT-Huge, and Semi-ViT outperforms the prior art such as SimCLRv2 \cite{DBLP:conf/nips/ChenKSNH20}. 3) Semi-ViT has shown the potential for a substantial reduction of labeling cost. For example, as seen in Figure~\ref{fig:sota}~(c), Semi-ViT-Huge with 1\% (10\%) ImageNet labels achieves a comparable performance of a fully-supervised Inception-v4 \cite{DBLP:conf/aaai/SzegedyIVA17} (ConvNeXt-L \cite{liu2022convnet}). This implies a $100\times$ ($10\times$) reduction in human annotation cost.
4) Semi-ViT achieves the state-of-the-art SSL results on ImageNet, \eg, 80.0\% (84.3\%) top-1 accuracy with only 1\% (10\%) labels. In addition, the substantial boost in performance by Semi-ViT is not isolated on ImageNet: we find an increase of 13\%-21\% (7\%-10\%) top-1 accuracy with 1\% (10\%) labels over the supervised fine-tuning baselines, for other datasets including Food-101 \cite{DBLP:conf/eccv/BossardGG14}, iNaturalist \cite{DBLP:journals/corr/HornASSAPB17} and GoogleLandmark \cite{DBLP:conf/iccv/NohASWH17}.

\section{Semi-supervised Vision Transformers}

\subsection{Pipeline}

Some pipelines for semi-supervised learning exist in the literature. For example:
1) The model is directly trained from scratch using SSL techniques, \eg, FixMatch \cite{DBLP:conf/nips/SohnBCZZRCKL20}; 
2) The model is un/self-supervised pretrained first and finetuned on the labeled data later \cite{DBLP:conf/cvpr/He0WXG20,DBLP:conf/icml/ChenK0H20,DBLP:conf/nips/GrillSATRBDPGAP20}; 
3) The model is self-supervised pretrained first and then semi-supervised finetuned on both labeled and unlabeled data \cite{DBLP:conf/cvpr/CaiRMFTS21}. 
In this paper, we instead present the following pipeline: at first, optional \emph{self-supervised pre-training} on all the data without using any labels; next, standard \emph{supervised fine-tuning} on the available labeled data; and finally, \emph{semi-supervised fine-tuning} on both labeled and unlabeled data. This procedure is similar to \cite{DBLP:conf/nips/ChenKSNH20}, with the difference that they use knowledge distillation \cite{DBLP:journals/corr/HintonVD15} in their final stage. We find that this training pipeline is stable to train semi-supervised vision transformers and achieves promising results, with possibly less hyperparameter tuning.

\begin{figure*}[t]
\begin{minipage}[b]{.49\linewidth}
\centering
\centerline{\epsfig{figure=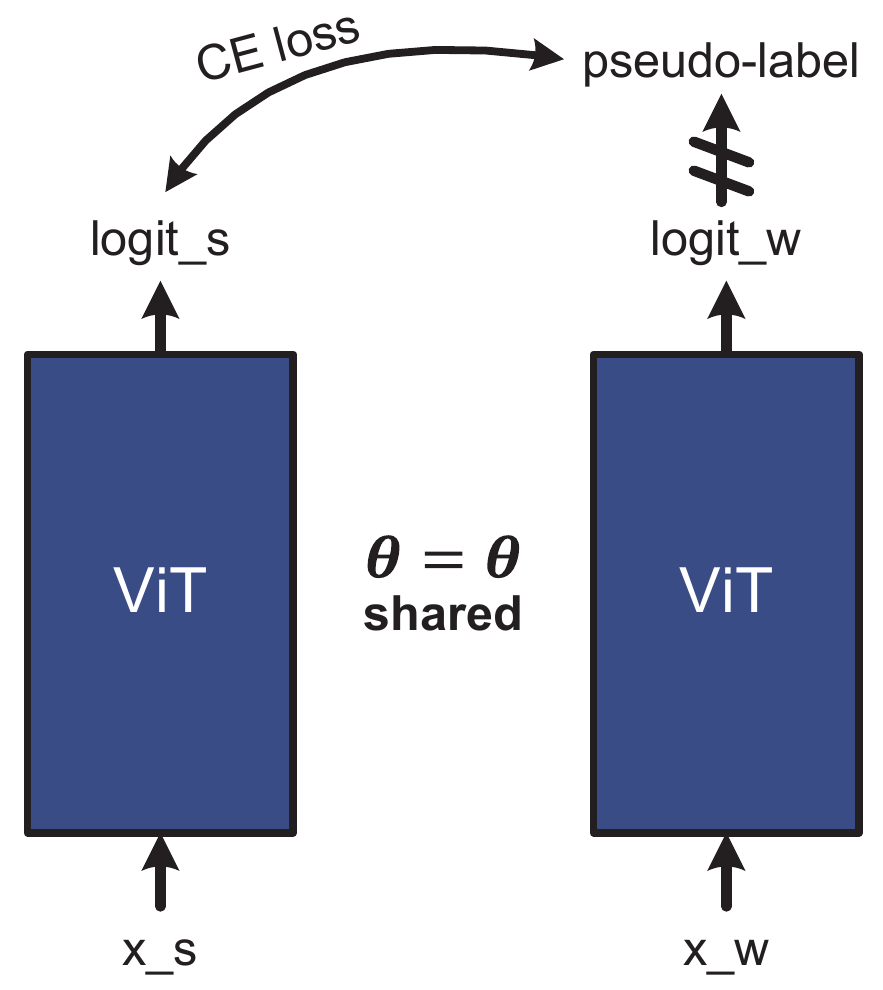,width=5cm,height=5.58cm}}{(a) FixMatch}
\end{minipage}
\hfill
\begin{minipage}[b]{.49\linewidth}
\centering
\centerline{\epsfig{figure=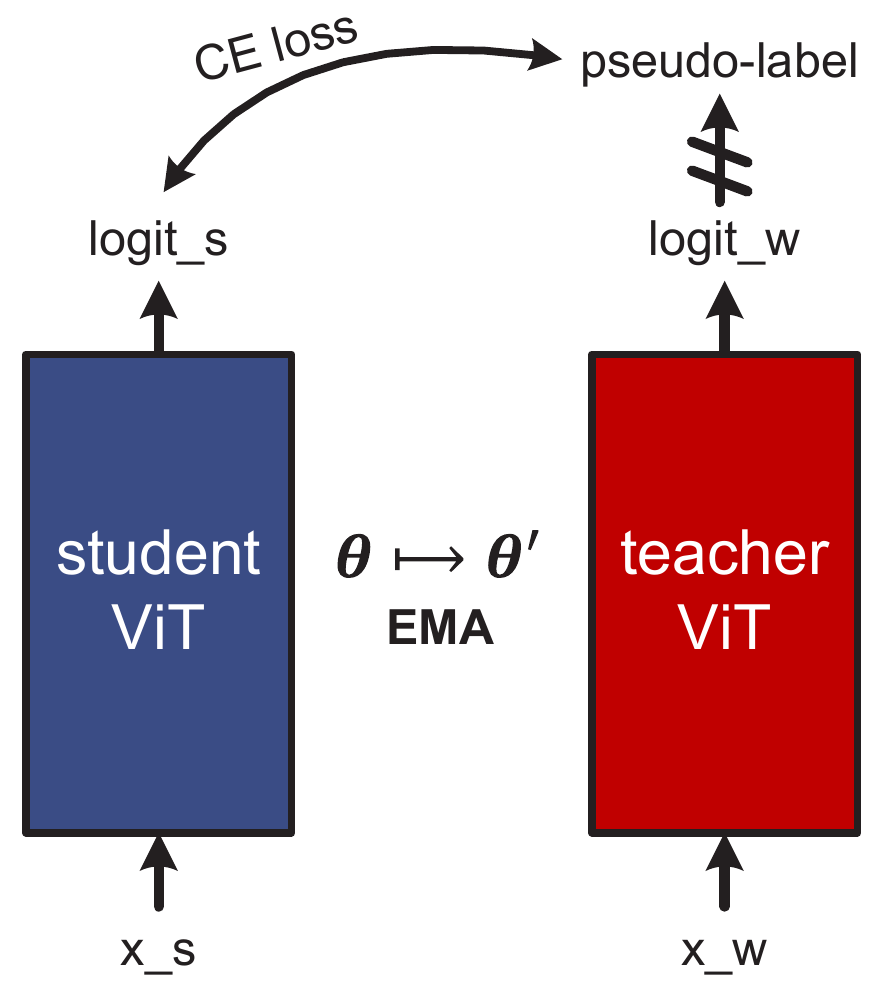,width=5cm,height=5.58cm}}{(b) EMA-Teacher}
\end{minipage}
\caption{The framework comparison between FixMatch (a) and EMA-Teacher (b). $x_s$/$x_w$ is the strongly/weakly augmented view of a sample $x$, and $\theta$ is the model parameters.}
\label{fig:framework}
\end{figure*}

\subsection{EMA-Teacher Framework}

FixMatch \cite{DBLP:conf/nips/SohnBCZZRCKL20} emerged as a popular SSL method in the past few years. As discussed in \cite{DBLP:conf/cvpr/CaiRMFTS21}, it can be interpreted as a student-teacher framework, where the student and teacher models are identical, as seen in Figure \ref{fig:framework} (a). 
However, FixMatch has unexpected behaviors, especially when the model consists of batch normalization (BN) \cite{DBLP:conf/icml/IoffeS15}. Although ViT uses Layer Normalization (LN) \cite{DBLP:journals/corr/BaKH16} instead of BN as normalization, we still found that FixMatch with ViT underperforms the CNN counterparts and often does not converge. This phenomenon was also observed in \cite{weng2021semi}. A potential reason to this is that the student and the teacher models are identical in FixMatch, which could easily lead to model collapse \cite{DBLP:conf/cvpr/He0WXG20,DBLP:conf/nips/GrillSATRBDPGAP20}. As suggested in \cite{DBLP:conf/cvpr/CaiRMFTS21}, the EMA-Teacher (shown in Figure \ref{fig:framework} (b)) is an improved version of FixMatch, thus we adopt it for our Semi-ViT. In the EMA-Teacher framework, the teacher parameters $\theta'$ are updated by exponential moving average (EMA) from the student parameters $\theta$, 
\begin{equation}
    \theta':=m\theta'+(1-m)\theta,
\label{equ:ema update}
\end{equation}
where the momentum decay $m$ is a number close to 1, \eg, 0.9999. The student parameters are updated by standard learning optimization, \eg, SGD or AdamW \cite{DBLP:conf/iclr/LoshchilovH19}. The other components are exactly the same as FixMatch, as seen in Figure \ref{fig:framework}. This temporal weight averaging can stabilize the training trajectories \cite{DBLP:conf/iclr/AthiwaratkunFIW19,DBLP:conf/uai/IzmailovPGVW18} and avoids the model collapse issue \cite{DBLP:conf/cvpr/He0WXG20,DBLP:conf/nips/GrillSATRBDPGAP20}. Our experiments also show this EMA-Teacher framework has better results and more stable training behaviors than FixMatch when training Semi-ViT.

\subsection{Semi-supervised Learning Formulation}

In the EMA-Teacher framework, there are both labeled and unlabeled
samples in a minibatch during training. 
The loss on the labeled samples $\{(x^l_i,y^l_i)\}^{N_l}_{i=1}$ is the standard cross-entropy loss, $\mathcal{L}_l=\frac{1}{N_l}\sum_{i=1}^{N_l}CE(x^l_i,y^l_i)$. For an unlabeled sample $x^u\in\{x^u_i\}^{N_u}_{i=1}$, a weak and a strong augmentation are applied to it, generating $x^{u,w}$ and $x^{u,s}$, respectively. The weak augmented $x^{u,w}$ is forwarded through the teacher network, and output the probabilities over classes, $p=f(x^{u,w};\theta')$. Then the pseudo label is produced by $\hat{y}=\arg\max_c p_c$ with its associated confidence $o=\max p_c$. The pseudo label with confidence higher than a confidence threshold $\tau$ is then used to supervise the learning of the student on the strong augmented sample $x^{u,s}$,
\begin{equation}
    \mathcal{L}_u=\frac{1}{N_u}\sum_{i=1}^{N_u}[o_i\geq\tau]CE(x^{u,s}_i,\hat{y}_i),
\label{equ:loss_u}
\end{equation}
where $[\cdot]$ is the indicator function. And the overall loss is $\mathcal{L}=\mathcal{L}_l+\mu\mathcal{L}_u$, where $\mu$ is the trade-off weight. Note that only the pseudo labels with confidence higher than a threshold contribute to the final loss; the others are instead not used. The philosophy behind this filtering is that the pseudo labels with low confidences are noisier and could hijack the SSL training.

\section{Probabilistic Pseudo Mixup}
\label{sec:mixup}

\subsection{Mixup}

Mixup \cite{DBLP:conf/iclr/ZhangCDL18} performs convex combinations of pairs of samples and their labels,
\begin{equation}
\begin{aligned}
    \tilde{x}&=\lambda x_i + (1-\lambda)x_j, \\
    \tilde{y}&=\lambda y_i + (1-\lambda)y_j,
\end{aligned}
\end{equation}
where the mixup ratio $\lambda\sim{Beta}(\alpha,\alpha)\in{[0,1]}$, for $\alpha\in(0,\infty)$. The samples are mixed-up usually in a single minibatch during training. Given a minibatch $\mathcal{B}$ and its shuffled version $\bar{\mathcal{B}}$, the mixed-up minibatch is $\tilde{\mathcal{B}}=\lambda\mathcal{B}+(1-\lambda)\bar{\mathcal{B}}$, where $\lambda$ could be either batch-wise or element-wise. Due to the nature of weak inductive bias, ViT is more data hungry than CNN, thus effective data augmentation, \eg, mixup, is critical for training fully-supervised ViT \cite{DBLP:conf/iclr/DosovitskiyB0WZ21,DBLP:conf/icml/TouvronCDMSJ21,DBLP:conf/iccv/LiuL00W0LG21,DBLP:conf/iccv/XuXCT21}. This also applies to Semi-ViT since it inherits the nature of weak inductive bias from ViT. Although it is standard to use mixup in supervised learning, how to employ it under pseudo-labeling based SSL framework, \eg, EMA-Teacher, is still unclear yet, and we are going to discuss it next.

\begin{figure*}[t!]
\begin{minipage}[b]{.32\linewidth}
\centering
\centerline{\includegraphics[width=1.0\linewidth]{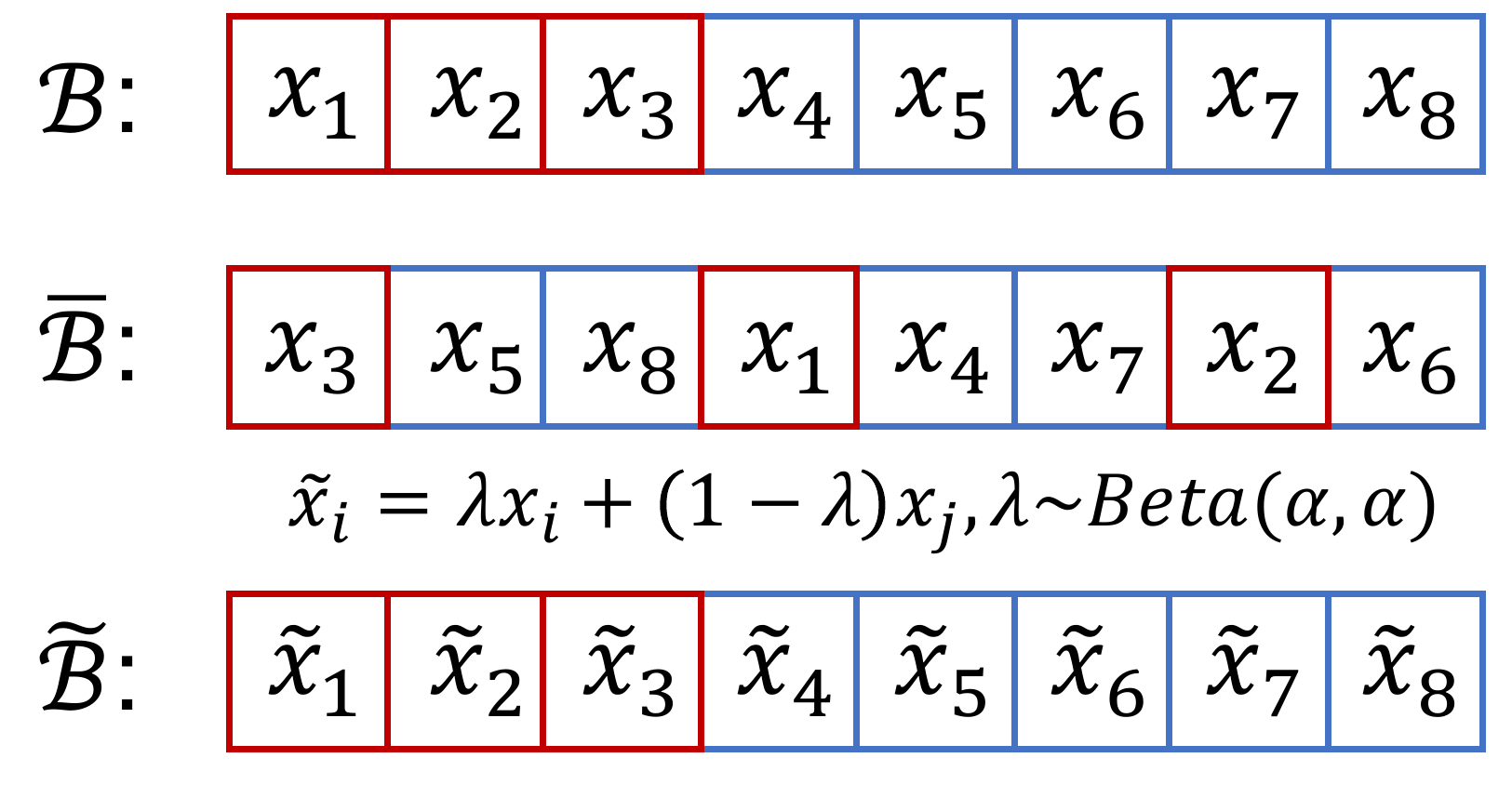}}{(a) Pseudo Mixup}
\end{minipage}
\hfill
\begin{minipage}[b]{.32\linewidth}
\centering
\centerline{\includegraphics[width=1.0\linewidth]{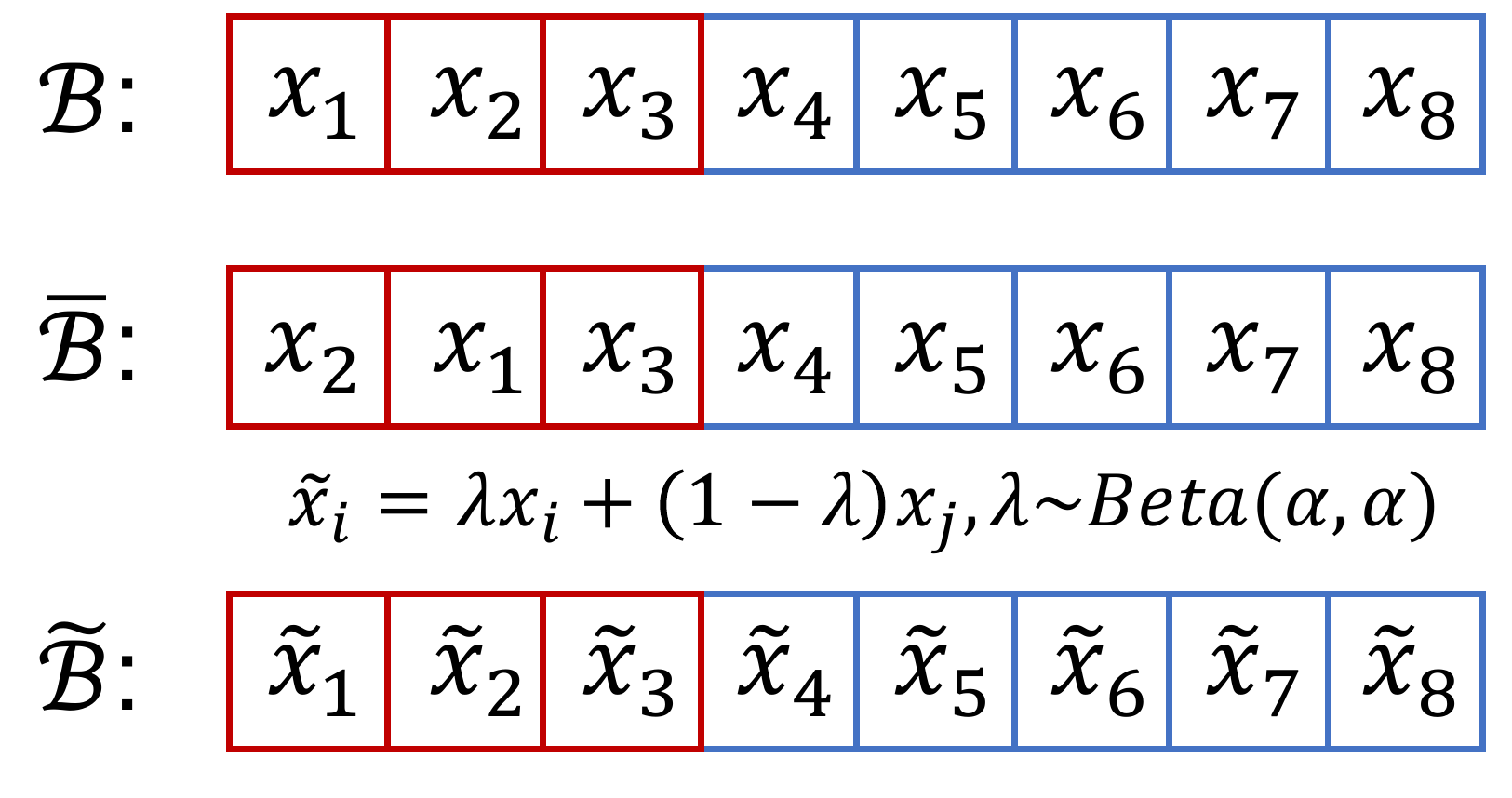}}{(b) Pseudo Mixup+}
\end{minipage}
\hfill
\begin{minipage}[b]{.32\linewidth}
\centering
\centerline{\includegraphics[width=1.0\linewidth]{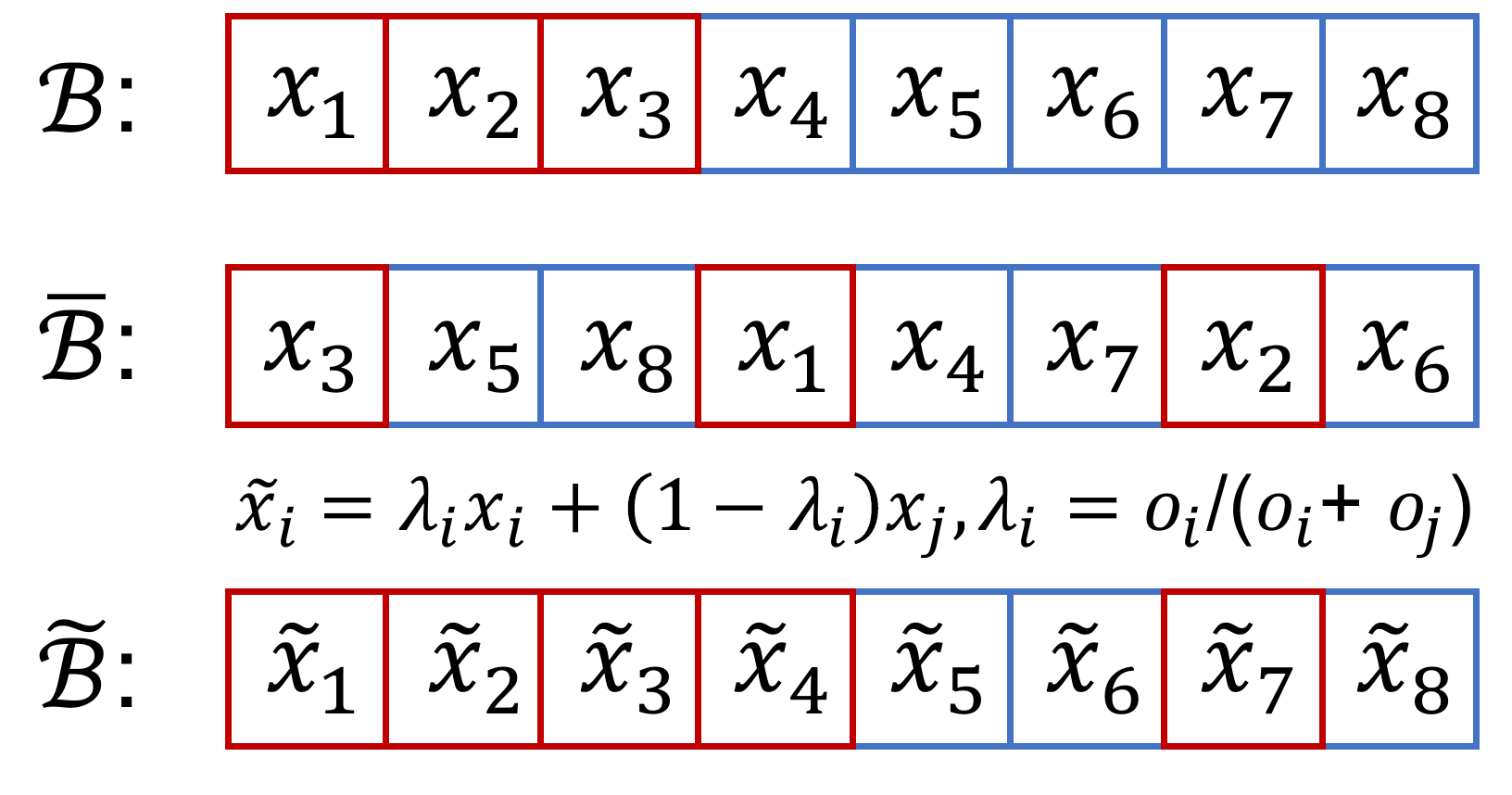}}{(c) Probabilistic Pseudo Mixup}
\end{minipage}
\caption{Different variations of mixup on unlabeled data. The red samples are the ones passing the confidence threshold, but not the blue samples.}
\label{fig:mixup}
\end{figure*}

\subsection{Pseudo Mixup}
\label{subse:pseudo mixup}

Under the pseudo-labeling based SSL framework \cite{lee2013pseudo,DBLP:conf/nips/SohnBCZZRCKL20,DBLP:conf/cvpr/PhamDXL21,DBLP:conf/cvpr/CaiRMFTS21}, given a unlabeled sample and its pseudo label $(x^u,\hat{y})$, only when its confidence $o$ is not smaller than the confidence threshold $\tau$, 
it will contribute to loss $\mathcal{L}_u$, as seen in (\ref{equ:loss_u}). According to their confidence scores, the unlabeled minibatch $\mathcal{B}^u$ can be grouped into a clean subset $\hat{\mathcal{B}}^u=\{(x^u_i,\hat{y}_i)|o_i\geq\tau\}$ and a noisy subset $\dot{\mathcal{B}}^u=\mathcal{B}^u-\hat{\mathcal{B}}^u$. 
One straightforward solution is to apply mixup on the full unlabeled minibatch $\mathcal{B}^u$, with no differentiation between clean and noisy samples, denoted as \emph{pseudo mixup}, as show in Figure \ref{fig:mixup} (a). After the pseudo mixup, still only the samples in $\hat{\mathcal{B}}^u$ contribute to the loss, and samples in $\dot{\mathcal{B}}^u$ are abandoned. In this way, the mixup operation is more than just a data augmentation. In fact, a sample in $\dot{\mathcal{B}}^u$ will also contribute to the final loss if it is mixed-up with a sample in $\hat{\mathcal{B}}^u$. As a result, it could involve a substantial number of noisy samples into the loss calculation due to the randomness, which, however, is against the philosophy of pseudo-labeling. Since only the clean subset $\hat{\mathcal{B}}^u$ contributes to the final loss, another choice is to use mixup only on $\hat{\mathcal{B}}^u$, denoted as \emph{pseudo mixup+}, as shown in Figure \ref{fig:mixup} (b). In this way, no sample in the noisy subset $\dot{\mathcal{B}}^u$ will affect the training.

\subsection{Probabilistic Pseudo Mixup}

Although the samples in $\dot{\mathcal{B}}^u$ are noisy, they still carry some useful information for the model to learn. The \emph{pseudo mixup} above can somehow leverage those information by blending the noisy and clean pseudo samples together. However, the problem is the mixup ratio is randomly generated from a Beta distribution, which does not depend on the confidence of each sample. This is not ideal. For example, when two samples are mixed-up, the sample with higher confidence should have higher mixup ratio, such that it can weigh more in the final loss. Motivated by this intuition, we propose \emph{probabilistic pseudo mixup} (Figure \ref{fig:mixup} (c)), where the mixup ratio $\lambda$ relfects the sample confidences,
\begin{equation}
    \lambda_i=o_i/(o_i+o_j).
\end{equation}
Also, the confidence score of $x^u_i$ is updated after mixup operation as
\begin{equation}
    o^*_i=\max(o_i,o_j),
\end{equation}
because the confidence score should align with the majority of the image content. And the final clean subset $\tilde{\mathcal{B}}^u=\{(\tilde{x}^u_i,\tilde{y}^u_i)|o^*_i\geq\tau\}$ will contribute to the final loss. Using this \emph{probabilistic pseudo mixup} can enhance regularization, leverage information from all samples, even the noisy ones, and not violate the philosophy of pseudo labeling at the same time. It can effectively alleviate the issue of weak inductive bias of Semi-ViT and bring substantial gains, as will be shown in our experiments.

\section{Experiments}

We evaluate Semi-ViT mainly on ImageNet, which consists of $\sim$1.28M training and 50K validation images. We sample 10\%/1\% labels from ImageNet training set for semi-supervised evaluation. 
We study both scenarios: with and without self-supervised pre-training. Without self-pretraining, we only evaluate on 10\% labels, since learning from scratch on 1\% labels is very difficult. 
When self-pretrained, MAE \cite{he2021masked} is mainly used, and we directly use their pretrained models. At the stage of \emph{supervised fine-tuning}, the model is trained for 100 (500) epochs with 5 (50) epochs of learning rate warmup with (without) self-supervised pre-training.
At the stage of \emph{semi-supervised fine-tuning}, the model is trained for 100 epochs, with 5 epochs of learning rate warmup. All learning is optimized with AdamW \cite{DBLP:conf/iclr/LoshchilovH19}, using cosine learning rate schedule, with a weight decay of 0.05. The momentum decay $m$ of (\ref{equ:ema update}) is 0.9999. In a minibatch, $N_u=5N_l$, and the loss trade-off $\mu=5$. The mixup is the combination of mixup \cite{DBLP:conf/iclr/ZhangCDL18} and Cutmix \cite{DBLP:conf/iccv/YunHCOYC19} as the implementation of \cite{rw2019timm}. More details can be found in the supplementary material.

\setlength{\tabcolsep}{8pt}
\begin{table}[t]
\begin{center}
\small
\begin{tabular}{lclccc}
\toprule
Model &Param & Method & 1\% & 10\% & 100\%\\\hline
\multirow{2}{*}{ViT-Base} &\multirow{2}{*}{86M} &finetune &57.4 &73.7 &83.7 \\
& &Semi-ViT &71.0 &79.7 &- \\\hline
\multirow{2}{*}{ViT-Large} &\multirow{2}{*}{307M} & finetune &67.1 &79.2 &86.0 \\
& &Semi-ViT &77.3 &83.3 &- \\\hline
\multirow{2}{*}{ViT-Huge} &\multirow{2}{*}{632M} & finetune &71.5 &81.4 &86.9 \\
& &Semi-ViT &80.0 &84.3 &- \\
\bottomrule
\end{tabular}
\caption{Semi-ViT results comparing with fine-tuning. The models are self-pretrained by MAE\cite{he2021masked}.}
\label{tab:semi-vit}
\end{center}\vspace{-3mm}
\end{table}

\subsection{Semi-ViT Results}

When the model is self-pretrained by MAE \cite{he2021masked}, we first evaluate the fine-tuning performances of MAE on the labeled data only, as the common practice in self/un-supervised learning literature \cite{DBLP:conf/cvpr/He0WXG20,DBLP:conf/icml/ChenK0H20,DBLP:conf/nips/GrillSATRBDPGAP20}, with results shown Table \ref{tab:semi-vit}. This already leads to strong semi-supervised baselines, \eg, 81.4 top-1 accuracy for ViT-Huge on 10\% labels, indicating MAE is a strong self-supervised learning technique. However, Semi-ViT has additional significant improvements over the strong baselines for all models, \eg, 8.5-13.6 points for 1\% labels and 2.9-6.0 points for 10\% labels. The fine-tuning results on 100\% data are provided as upper-bounds for our Semi-ViT, and their gaps to Semi-ViT are small, \eg, 4.0/2.7/2.6 points for ViT-Base/Large/Huge on 10\% labels. An interesting observation is that the larger model is more effective for smaller number of labels, which is consistent with the observations in \cite{DBLP:conf/nips/ChenKSNH20}. For example, the fine-tuning gaps between 1\% and 100\% labels are 26.3/18.9/15.4 points for ViT-Base/Large/Huge, which are decreasing. The observation on Semi-ViT results is similar, \eg, 12.7/8.7/6.9 points to their upper-bounds on 1\% labels. These results have shown that vision transformers can also perform very well in semi-supervised learning, as well as supervised learning and un/self-supervised learning.

\subsection{Ablation Studies}

\paragraph{FixMatch v.s. EMA-Teacher} is compared in Table~\ref{tab:fixmatch vs ema-teacher}. These experiments do not use the pseudo mixup techniques of Section~\ref{sec:mixup} yet. When the model is not self-pretrained, the training of FixMatch is unstable and often failed. When the model is self-pretrained, FixMatch training becomes stable, and start to achieve reasonable results, \eg, 74.8 for ViT-Base on 10\% labels, which is already better than the prior art on ResNet-50, \eg, 73.9 of MPL \cite{DBLP:conf/cvpr/PhamDXL21} and 74.0 of EMAN \cite{DBLP:conf/cvpr/CaiRMFTS21}. But it is only 1.7 points higher than the fine-tuning baseline of Table~\ref{tab:semi-vit}, indicating FixMatch is not an effective SSL framework for ViT. But EMA-Teacher achieves much better results, 3.3 points improvement over FixMatch when self-pretrained. Even without self-pretraining, EMA-Teacher can still achieve decent numbers, but FixMatch is failed.

\setlength{\tabcolsep}{8pt}
\begin{table}[t]
\begin{center}
\small
\begin{tabular}{lclcc}
\toprule
Model &Pretrained &Method & 1\% & 10\% \\\hline
\multirow{2}{*}{ViT-Small} &\multirow{2}{*}{None} &FixMatch &- &\xmark \\
& &EMA-Teacher &- &65.6 \\\hline
\multirow{2}{*}{ViT-Base} &\multirow{2}{*}{None} & FixMatch &- &\xmark \\
& &EMA-Teacher &- &68.9 \\\hline
\multirow{2}{*}{ViT-Base} &\multirow{2}{*}{MAE} & FixMatch &\xmark &74.8 \\
& &EMA-Teacher &65.3 &78.1\\
\bottomrule
\end{tabular}
\caption{The comparison between FixMatch and EMA-Teacher. \xmark  means the training is failed with accuracy close to 0.
}
\label{tab:fixmatch vs ema-teacher}
\end{center}\vspace{-3mm}
\end{table}

\setlength{\tabcolsep}{8pt}
\begin{table}[t]
\begin{center}
\small
\begin{tabular}{lclcc}
\toprule
Model &Pretrained &Mixup & 1\% & 10\% \\\hline
\multirow{4}{*}{ViT-Small} &\multirow{4}{*}{None} &EMA-Teacher &- &65.6 \\
& &Pseudo Mixup &- &68.3 \\
& &Pseudo Mixup+ &- &68.8 \\
& &ProbPseudo Mixup &- &70.9 \\\hline
\multirow{4}{*}{ViT-Base} &\multirow{4}{*}{None} &EMA-Teacher &- &68.9 \\
& &Pseudo Mixup &- &71.6 \\
& &Pseudo Mixup+ &- &72.1 \\
& &ProbPseudo Mixup &- &73.5 \\\hline
\multirow{4}{*}{ViT-Base} &\multirow{4}{*}{MAE} &EMA-Teacher &65.3 &78.1 \\
& &Pseudo Mixup &69.5 &78.3 \\
& &Pseudo Mixup+ &70.1 &78.7 \\
& &ProbPseudo Mixup &71.0 &79.7 \\
\bottomrule
\end{tabular}
\caption{The comparison among different mixup variations.
}
\label{tab:mixup}
\end{center}\vspace{-3mm}
\end{table}

\paragraph{Probabilistic Pseudo Mixup} Different mixup variations on unlabeled data are compared in Table \ref{tab:mixup}. Note that the standard mixup with implementation of \cite{rw2019timm} is used to the labeled data as usual. The EMA-Teacher does not use any mixup mechanism on the unlabeled data, serving as baselines here. When \emph{pseudo mixup} of Figure~\ref{fig:mixup} (a) is applied on the unlabeled data, the performances usually have some substantial gains over the EMA-Teacher baselines, especially for the scenarios that the training is more difficult, \eg, without self-pretraining or on 1\% labels. This shows the importance to use mixup on the unlabeled data for improved regularization. However, as discussed Section \ref{subse:pseudo mixup}, \emph{pseudo mixup} could involve many noisy samples into training. On the other hand, \emph{pseudo mixup+} of Figure~\ref{fig:mixup} (b) can increase over \emph{pseudo mixup} constantly, by about 0.5 points, showing that removing those noisy samples does help. In addition, \emph{probabilistic pseudo mixup} of Figure~\ref{fig:mixup} (c) can further improve over \emph{pseudo mixup+} by 1-2 points in all cases. These results have implied that those noisy samples do carry some useful information for SSL training, but their weights should be suppressed especially when their confidences are low. This data augmentation technique also effectively alleviate the training difficulty of semi-supervised vision transformers with weak inductive bias.

\paragraph{Effect of Self-pretraining} The self-pretraining of MAE \cite{he2021masked} has a substantial boost in performances, as seen in Table~\ref{tab:mixup}. For ViT-Base, MAE helps to improve by 6.2 and 9.2 points for EMA-Teacher with and without \emph{probabilistic pseudo mixup}, respectively. In addition, it helps to train the models in more challenging scenarios, \eg, 1\% labels. Without self-pretraining, the training fails to deliver good results on 1\% labels. Notice that, even without pre-training, our Semi-ViT (``ProbPseudo Mixup'' in Table \ref{tab:mixup}) also achieves slightly better performance than the CNN counterparts: 70.9 of Semi-ViT-Small v.s. 67.1 of FixMatch-ResNet50 or 69.2 \cite{DBLP:conf/nips/SohnBCZZRCKL20} of EMAN-ResNet50 \cite{DBLP:conf/cvpr/CaiRMFTS21} when trained from scratch for 100 epochs.

\setlength{\tabcolsep}{8pt}
\begin{table}[t]
\begin{center}
\small
\begin{tabular}{lllccc}
\toprule
Model &Pretrained & Method & 1\% & 10\% \\\hline
\multirow{3}{*}{ViT-Small} &\multirow{3}{*}{MoCo-v3 \cite{DBLP:conf/iccv/ChenXH21}} &finetune &51.2 &69.1 \\
& &EMA-Teacher &61.9 &72.3 \\
& &+ProbPseudo Mixup &64.7 &72.9 \\\hline
\multirow{3}{*}{ViT-Small} &\multirow{3}{*}{DINO \cite{DBLP:conf/iccv/CaronTMJMBJ21}} &finetune &58.7 &73.9 \\
& &EMA-Teacher &66.3 &76.3 \\
& &+ProbPseudo Mixup &68.0 &77.1 \\\hline
\multirow{3}{*}{ViT-Base} &\multirow{3}{*}{MoCo-v3 \cite{DBLP:conf/iccv/ChenXH21}} &finetune &66.3 &74.5 \\
& &EMA-Teacher &68.9 &77.7 \\
& &+ProbPseudo Mixup &72.3 &79.2 \\\hline
\multirow{3}{*}{ViT-Base} &\multirow{3}{*}{DINO \cite{DBLP:conf/iccv/CaronTMJMBJ21}} &finetune &65.0 &76.0 \\
& &EMA-Teacher &70.8 &78.1 \\
& &+ProbPseudo Mixup &73.1 &80.2 \\
\bottomrule
\end{tabular}
\caption{Semi-ViT results with other self-pretraining techniques.}
\label{tab:other self-pretraining}
\end{center}\vspace{-3mm}
\end{table}

\setlength{\tabcolsep}{8pt}
\begin{table}[t]
\begin{center}
\small
\begin{tabular}{lclc}
\toprule
Model & Upper-bound &Method & 10\% \\\hline
\multirow{3}{*}{ConvNeXt-T} &\multirow{3}{*}{80.7} & supervised & 61.2 \\
& &EMA-Teacher &70.4 \\
& &+ProbPseudo Mixup &74.1 \\\hline
\multirow{3}{*}{ConvNeXt-S} &\multirow{3}{*}{81.4} & supervised & 64.1 \\
& & EMA-Teacher &71.7 \\
& &+ProbPseudo Mixup &75.1 \\
\bottomrule
\end{tabular}
\caption{The results on ConvNeXt \cite{liu2022convnet}.}
\label{tab:convnext}
\end{center}\vspace{-3mm}
\end{table}

\paragraph{Other Self-pretraining Techniques}
Beyond MAE, we also experiment on other self-pretraining technique, including MoCo-v3 \cite{DBLP:conf/iccv/ChenXH21} and DINO \cite{DBLP:conf/iccv/CaronTMJMBJ21}, in Table~\ref{tab:other self-pretraining}. By comparing the fine-tuning results, DINO is close to MoCo-v3 for ViT-Base but much better for ViT-Small, and both of them are better than MAE for ViT-Base, suggesting that DINO could be a better self-pretraining technique for smaller scales of ViT models. On top of the strong fine-tuning baselines, the \emph{semi-supervised fine-tuning}, using EMA-Teacher, still has nontrivial improvements for both DINO and MoCo-v3, \eg, 5.8 (2.1) points on 1\% (10\%) labels for DINO-ViT-Base. In addition, the \emph{probabilistic pseudo mixup} can further improve over the EMA-Teacher, independent of the self-pretraining algorithms. And the final Semi-ViT-Base of DINO is 2.1 (0.5) points better than that of MAE on 1\% (10\%) labels.

\paragraph{Other Network Architectures}
Although in this paper we mainly focus on ViT architectures, the proposed \emph{probabilistic pseudo mixup} is not limited to them. We also try it for CNN architectures, \eg, ResNet \cite{DBLP:conf/cvpr/HeZRS16}. However, we find the direct use of the standard mixup does not improve fully-supervised ResNet performances, so will the \emph{probabilistic pseudo mixup} for its SSL setting. Instead, we evaluate it on the recently proposed ConvNeXt \cite{liu2022convnet}, which uses mixup for improved results. Since the goal is not to fully reproduce the results of \cite{liu2022convnet}, all models are trained only for 100 epochs, including the supervised upper-bounds. The results in Table \ref{tab:convnext} demonstrate that \emph{probabilistic pseudo mixup} is not limited to ViT, but also to CNN architectures, \eg, with improvements of 3-4 points, suggesting it can be well generalized.

\setlength{\tabcolsep}{10pt}
\begin{table}[t]
\begin{center}
\small
\begin{tabular}{lllccc}
\toprule
\multicolumn{2}{l}{Method} &Architecture &Param & 1\% & 10\% \\\hline
\parbox[t]{2mm}{\multirow{8}{*}{\rotatebox[origin=c]{90}{CNN}}} &UDA \cite{DBLP:conf/nips/XieDHL020} &ResNet-50 &26M &- &68.8 \\
&FixMatch \cite{DBLP:conf/nips/SohnBCZZRCKL20} &ResNet-50 &26M &- &71.5 \\
&S4L \cite{DBLP:conf/iccv/BeyerZOK19} &ResNet-50 (4$\times$) &375M &- &73.2 \\
&MPL \cite{DBLP:conf/cvpr/PhamDXL21} &ResNet-50 &26M &- &73.9 \\
&CowMix \cite{DBLP:conf/visapp/FrenchOS22} &ResNet-152 &60M &- &73.9 \\
&EMAN \cite{DBLP:conf/cvpr/CaiRMFTS21} &ResNet-50 &26M &63.0 &74.0 \\
&PAWS \cite{DBLP:conf/iccv/AssranCMBJBR21} &ResNet-50 &26M &66.5 &75.5 \\
&SimCLRv2+KD \cite{DBLP:conf/nips/ChenKSNH20} &RN152 (3$\times$+SK) &794M &76.6 &80.9 \\\hline
\parbox[t]{2mm}{\multirow{6}{*}{\rotatebox[origin=c]{90}{Transformer}}} &DINO \cite{DBLP:conf/iccv/CaronTMJMBJ21} &ViT-Small &22M &64.5 &72.2 \\
&SemiFormer \cite{weng2021semi} &ViT-S+Conv &42M &- &75.5 \\\cline{2-6}
&Semi-ViT (ours) &ViT-Small &22M &68.0 &77.1 \\
&Semi-ViT (ours) &ViT-Base &86M &71.0 &79.7 \\
&Semi-ViT (ours) &ViT-Large &307M &77.3 &83.3 \\
&Semi-ViT (ours) &ViT-Huge &632M &80.0 &84.3 \\
\bottomrule
\end{tabular}
\caption{The comparison with the state-of-the-art SSL models.}
\label{tab:sota}
\end{center}\vspace{-3mm}
\end{table}

\setlength{\tabcolsep}{10pt}
\begin{table}[t]
\begin{center}
\small
\begin{tabular}{llcccc}
\toprule
\multicolumn{2}{l}{Model} &Param &Data & top-1 & top-5 \\\hline
\parbox[t]{2mm}{\multirow{9}{*}{\rotatebox[origin=c]{90}{CNN}}} &ResNet-50 \cite{DBLP:conf/cvpr/HeZRS16} &26M &ImageNet &76.0 &93.0 \\
&ResNet-152 \cite{DBLP:conf/cvpr/HeZRS16} &60M &ImageNet &77.8 &93.8 \\
&DenseNet-264 \cite{DBLP:conf/cvpr/HuangLMW17} &34M &ImageNet &77.9 &93.9 \\
&Inception-v3 \cite{DBLP:conf/cvpr/SzegedyVISW16} &24M &ImageNet &78.8 &94.4 \\
&Inception-v4 \cite{DBLP:conf/aaai/SzegedyIVA17} &48M &ImageNet &80.0 &95.0 \\
&ResNeXt-101 \cite{DBLP:conf/cvpr/XieGDTH17} &84M &ImageNet &80.9 &95.6 \\
&SENet-154 \cite{DBLP:conf/cvpr/HuSS18} &146M &ImageNet &81.3 &95.5 \\
&ConvNeXt-L \cite{liu2022convnet} &198M &ImageNet &84.3 &- \\
&EfficientNet-L2 \cite{DBLP:conf/icml/TanL19} &480M &ImageNet &85.5 &97.5 \\\hline
\parbox[t]{2mm}{\multirow{6}{*}{\rotatebox[origin=c]{90}{Transformer}}} &ViT-Huge \cite{DBLP:conf/iclr/DosovitskiyB0WZ21} &632M &JFT+ImageNet &88.6 &- \\
&DeiT-B \cite{DBLP:conf/icml/TouvronCDMSJ21}  &86M &ImageNet &81.8 &- \\
&Swin-B \cite{DBLP:conf/iccv/LiuL00W0LG21} &88M &ImageNet &83.3 &- \\
&MAE-ViT-Huge \cite{he2021masked} &632M &ImageNet &86.9 &- \\\cline{2-6}
&Semi-ViT-Huge (ours) &632M &1\%ImageNet &80.0 &93.1 \\
&Semi-ViT-Huge (ours) &632M &10\%ImageNet &84.3 &96.6 \\
\bottomrule
\end{tabular}
\caption{The comparison with the state-of-the-art fully supervised models.}
\label{tab:full sota}
\end{center}\vspace{-3mm}
\end{table}

\subsection{Comparison with the State-of-the-Art}

Semi-ViTs are compared with the state-of-the-art semi-supervised learning algorithms in Table \ref{tab:sota}. When the model capacity is close, our Semi-ViT has shown much better results than the prior art, \eg, MPL-RN-50 \cite{DBLP:conf/cvpr/PhamDXL21} v.s. Semi-ViT-Small, CowMix-RN152 \cite{DBLP:conf/visapp/FrenchOS22} v.s. Semi-ViT-Base, S4L-RN50-4$\times$ \cite{DBLP:conf/iccv/BeyerZOK19} v.s. Semi-ViT-Large and SimCLRv2+KD-RN152-3$\times$-SK \cite{DBLP:conf/nips/ChenKSNH20} v.s. Semi-ViT-Huge. The only transformer based SSL method is SemiFormer \cite{weng2021semi}, but it requires to use CNN as the teacher model and blend convolution and transformer modules together for good performances. However, our Semi-ViT is \emph{pure} ViT based, without any additional parameters and architecture changes, and the Semi-ViT-Small model is already better than SemiFormer (77.1 v.s. 75.5). These comparisons support that Semi-ViT does advance the state-of-the-art of semi-supervised learning.

Scalability is an advantage of ViT, and we compare the scalability of Semi-ViT with previous works in Figure \ref{fig:sota} (a) and (b). The comparison has shown that Semi-ViT can achieve better trade-off between model capacity and accuracy and can be scaled up more effectively than the prior art, SimCLRv2 \cite{DBLP:conf/nips/ChenKSNH20}. For example, SimCLRv2 and PAWS \cite{DBLP:conf/iccv/AssranCMBJBR21} scale up the model usually in terms of network depth and width, and they seem to saturate when the model is of medium size, \eg, around 300M parameters, but our Semi-ViT continues to improve steadily beyond that point. 

Semi-ViT is also compared with the supervised state-of-the-art in Table \ref{tab:full sota}. Our Semi-ViT-Huge is comparable with Inception-v4 \cite{DBLP:conf/aaai/SzegedyIVA17}, but with 100$\times$ annotation cost reduction, and comparable with ConvNeXt-L \cite{liu2022convnet} (better than Swin-B \cite{DBLP:conf/iccv/LiuL00W0LG21}), but with 10$\times$ annotation cost reduction. These comparisons imply that Semi-ViT has great potential for labeling cost reduction.

\setlength{\tabcolsep}{8pt}
\begin{table}[t]
\begin{center}
\small
\begin{tabular}{lcclccc}
\toprule
Dataset &\# train/test &\# class &Method & 1\% & 10\% & 100\% \\\hline
\multirow{2}{*}{Food-101 \cite{DBLP:conf/eccv/BossardGG14}} &\multirow{2}{*}{75.7K/25.2K} &\multirow{2}{*}{101} &Finetune &60.9 &84.5 &93.1 \\
& & &Semi-ViT &82.1 &91.3 &- \\\hline
\multirow{2}{*}{iNaturalist \cite{DBLP:journals/corr/HornASSAPB17}} &\multirow{2}{*}{265K/3K} &\multirow{2}{*}{1010} & Finetune &19.6 &57.3 &81.2 \\
& & &Semi-ViT &32.3 &67.7 &- \\\hline
\multirow{2}{*}{GoogleLandmark \cite{DBLP:conf/iccv/NohASWH17}} &\multirow{2}{*}{200K/15.6K} &\multirow{2}{*}{256} & Finetune &45.3 &74.0 &91.5 \\
& & &Semi-ViT &61.0 &81.0 &- \\
\bottomrule
\end{tabular}
\caption{The Semi-ViT-Base results on other datasets.}
\label{tab:other datasets}
\end{center}\vspace{-3mm}
\end{table}

\subsection{Other Datasets}

The generalization of Semi-ViT is evaluated on datasets including Food-101 \cite{DBLP:conf/eccv/BossardGG14}, iNaturalist \cite{DBLP:journals/corr/HornASSAPB17} and GoogleLandmark \cite{DBLP:conf/iccv/NohASWH17}. Since these datasets are beyond ImageNet, we assume that the ImageNet dataset is available and the model is already supervised pretrained on ImageNet, and then the model is finetuned to different target datasets with a few labels. The results are shown in Table \ref{tab:other datasets}. On these dataset, our Semi-ViT can improve over the fine-tuning baselines by 13-21 (7-10) points on 1\% (10\%) labels. Note that on Food-101, Semi-ViT on 1\% (10\%) labels is close to fine-tuning baseline on 10\% (100\%) labels, \ie, 82.1 v.s. 84.5 (91.3 v.s. 93.1), indicating that using Semi-ViT can help to save annotation costs by about 10 times on this dataset.

\section{Related Work}

Semi-supervised learning has a long history of research \cite{zhu2005semi,chapelle2009semi}. The recent works can be roughly clustered into two groups, consistency-based \cite{DBLP:conf/iclr/LaineA17,DBLP:conf/nips/TarvainenV17,DBLP:journals/pami/MiyatoMKI19,DBLP:conf/nips/XieDHL020,DBLP:conf/ijcai/VermaLKBL19} and pseudo-labeling based \cite{lee2013pseudo,DBLP:conf/nips/SohnBCZZRCKL20,DBLP:conf/cvpr/PhamDXL21,DBLP:conf/cvpr/CaiRMFTS21}. Consistency-based methods usually add some noise to the input or model, and then enforce their feature or probability outputs to be consistent. For example, to construct two outputs for later consistency regularization, $\Pi$-model \cite{DBLP:conf/iclr/LaineA17} adds noise to the model weights using dropout \cite{DBLP:journals/jmlr/SrivastavaHKSS14}, Mean-teacher \cite{DBLP:conf/nips/TarvainenV17} builds a teacher model by EMA updated from the student model, and UDA \cite{DBLP:conf/nips/XieDHL020} applies a weak and a strong data augmentation to the input. On the other hand, the idea of pseudo-labeling or self-training can be traced back to \cite{DBLP:journals/tit/Scudder65a,mclachlan1975iterative}, which uses model predictions as hard pseudo labels to guide the learning on unlabeled data. This idea becomes popular in SSL recently \cite{lee2013pseudo,DBLP:conf/nips/SohnBCZZRCKL20,DBLP:conf/cvpr/PhamDXL21,DBLP:conf/cvpr/CaiRMFTS21,DBLP:conf/cvpr/XieLHL20}. In the offline pseudo labeling \cite{lee2013pseudo,DBLP:conf/cvpr/XieLHL20}, the model used to generate pseudo labels is usually frozen or updated once in a while during training, \eg, at the end of every training epoch, but for online pseudo-labeling \cite{DBLP:conf/nips/SohnBCZZRCKL20,DBLP:conf/cvpr/CaiRMFTS21} the teacher model is updated continuously along with the student. Beyond classification, pseudo-labeling has also achieved promising progresses in more challenging tasks, \eg, object detection \cite{sohn2020simple,DBLP:conf/iclr/LiuMHKCZWKV21,wang22omni}. Our Semi-ViT also falls into the category of online pseudo-labeling. 

Mixup \cite{DBLP:conf/iclr/ZhangCDL18} is an effective data augmentation technique, which interpolates the input samples and their labels linearly and performs vicinal risk minimization. It has been successfully used in image classification and some other domains, \eg, generative adversarial networks \cite{DBLP:conf/icml/LucasTOV18}, sentence classification \cite{DBLP:conf/aaai/Guo20}, etc. Other variants have also been developed, \eg, Manifold Mixup \cite{DBLP:conf/icml/VermaLBNMLB19} that mixes up in the feature space or CutMix \cite{DBLP:conf/iccv/YunHCOYC19} which cuts a patch from one image and pastes it into another one. Mixup has also been successfully adopted in self-supervised learning \cite{DBLP:conf/nips/KalantidisSPWL20,DBLP:conf/iclr/LeeZSLSL21} and semi-supervised learning \cite{DBLP:conf/nips/BerthelotCGPOR19,DBLP:conf/ijcai/VermaLKBL19,DBLP:conf/iclr/BerthelotCCKSZR20}. Although \cite{DBLP:conf/nips/BerthelotCGPOR19,DBLP:conf/ijcai/VermaLKBL19,DBLP:conf/iclr/BerthelotCCKSZR20} also used mixup for SSL, they have differences with our \emph{probabilistic pseudo mixup}: 1) they are consistency-based SSL framework, but ours is pseudo-labeling based; 2) their mixup ratio is random sampled, but ours depends on the pseudo label confidence; 3) they have only shown successes on small CNN architectures and small datasets, \eg, CIFAR \cite{krizhevsky2009learning} and SVHN \cite{netzer2011reading}, but our successes are built on various scales of transformer architectures and large-scale datasets, \eg, ImageNet \cite{DBLP:journals/ijcv/RussakovskyDSKS15}, INaturalist \cite{DBLP:journals/corr/HornASSAPB17}, GoogleLandmark \cite{DBLP:conf/iccv/NohASWH17}, etc.

\section{Conclusion}

In this paper, we propose Semi-ViT for vision transformers based semi-supervised learning. This is the first time that \emph{pure} vision transformers can achieve promising results on semi-supervised learning and even surpass the previous best CNN based counterparts by a large margin. In addition, Semi-ViT inherits the scalable benefits from ViT, and the larger model leads to smaller gap to the fully supervised upper-bounds. This has shown to be a promising direction for semi-supervised learning. And the advantages of Semi-ViT can be well generalized to other datasets, suggesting potentially broader impacts. We hope these promising results could encourage more efforts in semi-supervised vision transformers.

{
\small
\bibliographystyle{plain}
\bibliography{egbib}
}


\newpage
\appendix

\section{Implementation Details}

\paragraph{Dataset Sampling} We sample 1\% (10\%) images per class from the datasets we use for the semi-supervised learning experiments of 1\% (10\%) labels. For example, on ImageNet, the number of sampled images is 12,820 (128,118) in total, for 1\% (10\%) labels.

\paragraph{ViT Architectures} We use exactly same architecture as the standard ViT \cite{DBLP:conf/iclr/DosovitskiyB0WZ21}. For the position tokens, we use the learnable positional embedding when the model is self-pretrained, but the sine-cosine version of positional embedding (non-learnable) when not self-pretrained since we find it leads to better results than the learnable one. For all architectures, the classifier is built on top of the average pooling of the encoder output, except for ViT-Huge where we find the classifier on top of the output of the class token leads to higher performances.

\paragraph{Data Augmentation}
We use the common data augmentations of \texttt{RandomResizedCrop}, \texttt{RandomHorizontalFlip}, \texttt{RandAugment(`m9-mstd0.5-inc1')} \cite{DBLP:conf/nips/CubukZS020} and \texttt{RandomErasing} \cite{DBLP:conf/aaai/Zhong0KL020} on the labeled data. The same augmentations are also used on the unlabeled data as the strong augmentation in EMA-Teacher, whereas the weak augmentations are  \texttt{RandomResizedCrop}, \texttt{RandomHorizontalFlip} and \texttt{ColorJitter(0.4)}. We do not extensively explore the space of data augmentation for weak and strong augmentations. The center 224$\times$224 crop is used at inference.

\paragraph{Self-supervised Pretraining} We directly use the pretrained models from MAE \cite{he2021masked}, DINO \cite{DBLP:conf/iccv/ChenXH21} and MoCo-v3 \cite{DBLP:conf/iccv/CaronTMJMBJ21}. The final Semi-ViT-Small is with DINO self-pretraining.

\paragraph{Supervised Fine-tuning Settings}  
The settings for the stage of \emph{supervised fine-tuning}, with and without self-pretraining, are shown in Table \ref{tab:impl_finetune}. These settings are mainly following the settings of finetuning and learning from scratch in \cite{he2021masked}, with some minor modifications. We also use the linear learning rate scaling rule \cite{DBLP:journals/corr/GoyalDGNWKTJH17}: $lr=base\_lr\times{batchsize}/256$. When supervised fine-tuning on 1\% data, we find the performances are bad when the regularization is strong, hence we do not use mixup/cutmix, drop path and random erasing, in that case, except ViT-Small. 

\paragraph{Semi-supervised Fine-tuning Settings}
The settings for the stage of \emph{semi-supervised fine-tuning}, with and without self-pretraining, are shown in Table \ref{tab:impl_ssl}. When semi-supervised fine-tuning small models (ViT-Small/Base) on 1\% data, we find the performances are bad when the regularization is strong, hence we do not use mixup/cutmix on labeled data, and drop path and random erasing on both labeled and unlabeled data, in that case. The linear learning rate scaling rule is: $lr=base\_lr\times{batchsize_l}/256$, where $batchsize_l$ is the batch size of the labeled data.

\paragraph{Confidence Threshold} The confidence thresholds of Semi-ViT are shown in Table \ref{tab:impl_ssl}. But those are not optimal for FixMatch/EMA-Teacher experiments without \emph{probabilistic pseudo mixup}, and instead, we search the optimal $\tau$ from $\{0.5,0.6,0.7,0.8,0.9\}$ for them.

\paragraph{Computing Resources} We run all experiments on V100 GPUs of 32G memory. For Semi-ViT-Small/Base/Large/Huge, we use 8/8/16/32 GPUs for training 100/100/100/50 epochs, which takes about 19/31/61/115 hours.

\setlength{\tabcolsep}{10pt}
\begin{table}[t]
\begin{center}
\small
\begin{tabular}{l|lll}
config & 10\% labels  & 1\% labels & 10\% labels ( scratch) \\
\toprule
optimizer & AdamW & AdamW & AdamW  \\
\multirow{2}{*}{base learning rate} & 1e-4 (S), 2.5e-4 (B) & 1e-4 (S), 5e-5 (B) & \multirow{2}{*}{1e-4} \\
& 1e-3 (L/H) & 1e-3 (L), 0.01 (H) & \\
weight decay & 0.05 & 0.05 & 0.3 \\
optimizer momentum & $\beta_1, \beta_2{=}0.9, 0.999$ & $\beta_1, \beta_2{=}0.9, 0.999$ & $\beta_1, \beta_2{=}0.9, 0.95$ \\
layer-wise lr decay \cite{bao2021beit} & 0.65 (S/B), 0.75 (L/H) & 0.65 (S/B), 0.75 (L/H) & 1.0 \\
batch size & 512 (S/B/L), 256 (H) & 512 (S/B/L), 128 (H) & 1024 \\
learning rate schedule & cosine decay & cosine decay & cosine decay \\
warmup epochs & 5  & 5 & 50 \\
training epochs & 100 (S/B), 50 (L/H) & 100 (S/B), 50 (L/H) & 500 \\
label smoothing \cite{DBLP:conf/cvpr/SzegedyVISW16} & 0.1 & 0.1 & 0.1 \\
mixup \cite{DBLP:conf/iclr/ZhangCDL18} & 0.8 &  0.8 (S), 0 (B/L/H) & 0.8 \\
cutmix \cite{DBLP:conf/iccv/YunHCOYC19} & 1.0 & 1.0 (S), 0 (B/L/H) & 1.0 \\
drop path \cite{DBLP:conf/eccv/HuangSLSW16} & 0.1 (S/B/L) 0.2 (H) & 0.1 (S), 0 (B/L/H) & 0.1 \\
random erasing \cite{DBLP:conf/aaai/Zhong0KL020} & 0.25 & 0.25 (S), 0 (B/L/H) & 0.25 \\
\end{tabular}
\caption{Supervised fine-tuning settings with and without self-pretraining.}
\label{tab:impl_finetune}\vspace{-3mm}
\end{center}
\end{table}

\setlength{\tabcolsep}{10pt}
\begin{table}[t]
\begin{center}
\small
\begin{tabular}{l|lll}
config & 10\% labels  & 1\% labels & 10\% labels ( scratch) \\
\toprule
optimizer & AdamW & AdamW & AdamW  \\
\multirow{2}{*}{base learning rate} & 2e-4 (S), 1e-3 (B) & 5e-4 (S), 1e-3 (B/L) & \multirow{2}{*}{1e-3} \\
& 2e-3 (L), 2.5e-3 (H) & 5e-3 (H) & \\
weight decay & 0.05 & 0.05 & 0.05 \\
optimizer momentum & $\beta_1, \beta_2{=}0.9, 0.999$ & $\beta_1, \beta_2{=}0.9, 0.999$ & $\beta_1, \beta_2{=}0.9, 0.999$ \\
layer-wise lr decay \cite{bao2021beit} & 0.75 & 0.75 & 0.85 \\
batch size (labeled) & 128 & 128 (S/B/L), 64 (H) & 128 \\
learning rate schedule & cosine decay & cosine decay & cosine decay \\
confidence threshold & 0.5 (S/B), 0.6 (L/H) & 0.6 & 0.5 \\
warmup epochs & 5  & 5 & 5 \\
training epochs & 100 (S/B/L), 50 (H) & 100 (S/B/L), 50 (H) & 100 \\
label smoothing \cite{DBLP:conf/cvpr/SzegedyVISW16} & 0.1 & 0.1 & 0.1 \\
mixup \cite{DBLP:conf/iclr/ZhangCDL18} & 0.8 & 0.8 & 0.8 \\
cutmix \cite{DBLP:conf/iccv/YunHCOYC19} & 1.0 & 1.0 & 1.0 \\
drop path \cite{DBLP:conf/eccv/HuangSLSW16} & 0.1 (S/B/H), 0.2 (L) & 0 (S/B), 0.1 (L), 0.05 (H) & 0.1 \\
random erasing \cite{DBLP:conf/aaai/Zhong0KL020} & 0.25 & 0 (S/B), 0.25 (L/H) & 0.25 \\
\end{tabular}
\caption{Semi-supervised fine-tuning settings with and without self-pretraining.}
\label{tab:impl_ssl}\vspace{-3mm}
\end{center}
\end{table}

\paragraph{Random Seeds and Error Bar}
Since some of the experiments are expensive to run, e.g., Semi-ViT-Huge. We only test the randomness on Semi-ViT-Base models. We sample three different 10\%/1\% subsets of ImageNet and repeat the experiments for three times with different random seeds. When self-pretrained by MAE, the accuracy is $79.71\pm{0.037}$ ($70.95\pm{0.029}$) for 10\% (1\%) labels. When not self-pretrained, the accuracy is $73.44\pm{0.065}$ for 10\% labels. The results have shown that Semi-ViT is quite robust to different random seeds and different subsets of the samples. To keep consistency, all experiments in the paper are run with the same ImageNet subset and the same random seed.

\paragraph{Settings on Other Datasets}
The model is ViT-Base on the datasets of Food-101 \cite{DBLP:conf/eccv/BossardGG14}, iNaturalist \cite{DBLP:journals/corr/HornASSAPB17} and GoogleLandmark \cite{DBLP:conf/iccv/NohASWH17}. The \emph{supervised fine-tuning} settings are almost the same as those with self-pretraining in Table \ref{tab:impl_finetune}, with the differences: 1) we search the optimal base learning rate from $\{1e^{-4}, 2.5e^{-4}, 5e^{-4}, 1e^{-3}\}$ and the optimal layer-wise learning rate decay from $\{0.65, 0.75\}$; 2) we enable mixup/cutmix, drop path and random erasing for 1\% experiments (except iNaturalist), with the same values of those for 10\% experiments. The \emph{semi-supervised fine-tuning} settings are almost the same as those with self-pretraining in Table \ref{tab:impl_ssl}, with the difference that we set the EMA momentum decay $m=0.999$ instead of $m=0.9999$ as in the ImageNet experiments, since these datasets are smaller and need faster EMA update rate.



\end{document}